\documentclass[lettersize,journal]{IEEEtran}
\usepackage{amsmath,amsfonts}
\usepackage{algorithmic}
\usepackage{algorithm}
\usepackage{array}
\usepackage[caption=false,font=normalsize,labelfont=sf,textfont=sf]{subfig}
\usepackage{textcomp}
\usepackage{stfloats}
\usepackage{url}
\usepackage{verbatim}
\usepackage{graphicx}
\usepackage{cite}
\usepackage{booktabs}
\usepackage{tabularx}
\usepackage{multirow}
\usepackage{arydshln}
\usepackage{ragged2e}
\usepackage{adjustbox}
\usepackage{amsmath,amssymb,amsfonts}
\usepackage{cite}
\usepackage{graphicx}
\usepackage{xcolor}
\usepackage{xpatch}
 \usepackage{array}
\usepackage[colorlinks,bookmarksopen,bookmarksnumbered,citecolor=blue, linkcolor=blue, urlcolor=blue]{hyperref}
\begin{document}

\title{Ultrasound Nodule Segmentation Using Asymmetric Learning with Simple Clinical Annotation}

\author{Xingyue Zhao, Zhongyu Li, Xiangde Luo, Peiqi Li, Peng Huang, Jianwei Zhu, Yang Liu, Jihua Zhu, Meng Yang, Shi Chang, Jun Dong
\thanks{X. Zhao, Z. Li and X. Luo contributed equally to this work. Corresponding authors: Z. Li (zhongyuli@xjtu.edu.cn) and J. Dong (dongjun\_orth@xjtu.edu.cn).}
\thanks{Xingyue Zhao, Zhongyu Li, Peiqi Li, Jianwei Zhu and Jihua Zhu are with School of Software Engineering, Xi’an Jiaotong University,
Xi’an, China, also with Shaanxi Joint Key Laboratory for Artifact Intelligence, China.
}
\thanks{Xiangde Luo is with the School of Mechanical and Electrical Engineering, University of Electronic Science and Technology of China, Chengdu 610072, China and is also with Shanghai Artificial Intelligence Laboratory, Shanghai 200030, China.}
\thanks{Shi Chang and Peng Huang are with Department of General Surgery, Xiangya Hospital, Central South University, China.
}
\thanks{Yang Liu is with School of Cyber Science and Engineering, MoE KLINNS Lab, Xi’an Jiaotong University, Xi’an 710049, China.}
\thanks{Jun Dong is with Department of Orthopaedics, Second Affiliated Hospital of Xi'an Jiaotong University, Xi'an 710004, China.}
\thanks{Meng Yang is with Hunan Frontline Medical Technology Co., Ltd.}
\thanks{This work is supported by the Key Research and Development Program of Shaanxi Province under Grant. 2021GXLH-Z-097, 2024SF-YBXM-196.}
}

\markboth{Copyright © 20xx IEEE. Personal use of this material is permitted. However, permission to use this material for any other purposes must be obtained from the IEEE by sending an email to pubs-permissions@ieee.org.}%
{Shell \MakeLowercase{\textit{et al.}}: bare Demo of IEEEtran.cls for IEEE Journals}

\maketitle

\begin{abstract}
Recent advances in deep learning have greatly facilitated the automated segmentation of ultrasound images, which is essential for nodule morphological analysis. Nevertheless, most existing methods depend on extensive and precise annotations by domain experts, which are labor-intensive and time-consuming. In this study, we suggest using simple aspect ratio annotations directly from ultrasound clinical diagnoses for automated nodule segmentation. Especially, an asymmetric learning framework is developed by extending the aspect ratio annotations with two types of pseudo labels, i.e., conservative labels and radical labels, to train two asymmetric segmentation networks simultaneously. Subsequently, a conservative-radical-balance strategy (CRBS) strategy is proposed to complementally combine radical and conservative labels. An inconsistency-aware dynamically mixed pseudo-labels supervision (IDMPS) module is introduced to address the challenges of over-segmentation and under-segmentation caused by the two types of labels. To further leverage the spatial prior knowledge provided by clinical annotations, we also present a novel loss function namely the clinical anatomy prior loss. Extensive experiments on two clinically collected ultrasound datasets (thyroid and breast) demonstrate the superior performance of our proposed method, which can achieve comparable and even better performance than fully supervised methods using ground truth annotations.
\end{abstract}

\begin{IEEEkeywords}
ultrasound nodule segmentation, weakly supervised segmentation, aspect ratio annotations.
\end{IEEEkeywords}

\section{Introduction}
\label{intro}
\IEEEPARstart{D}{eep} learning has achieved promising advancements in the realm of automatic medical image segmentation, with most deep learning-based techniques requiring a large number of training images accompanied by accurate pixel-wise annotations. The Segment Anything Model (SAM) \cite{kirillov2023segment}, introduced in recent work, has brought a significant change in natural image segmentation. However, its effectiveness in medical imaging is limited due to the difficulties involved in collecting a large-scale, meticulously annotated medical image dataset. The process remains labor-intensive and time-consuming, requiring significant domain expertise and a deep understanding of clinical practices.

 \begin{figure}[!tb]
    \setlength\tabcolsep{1.6pt}
    \centering
    \small
    \includegraphics[width=\linewidth]{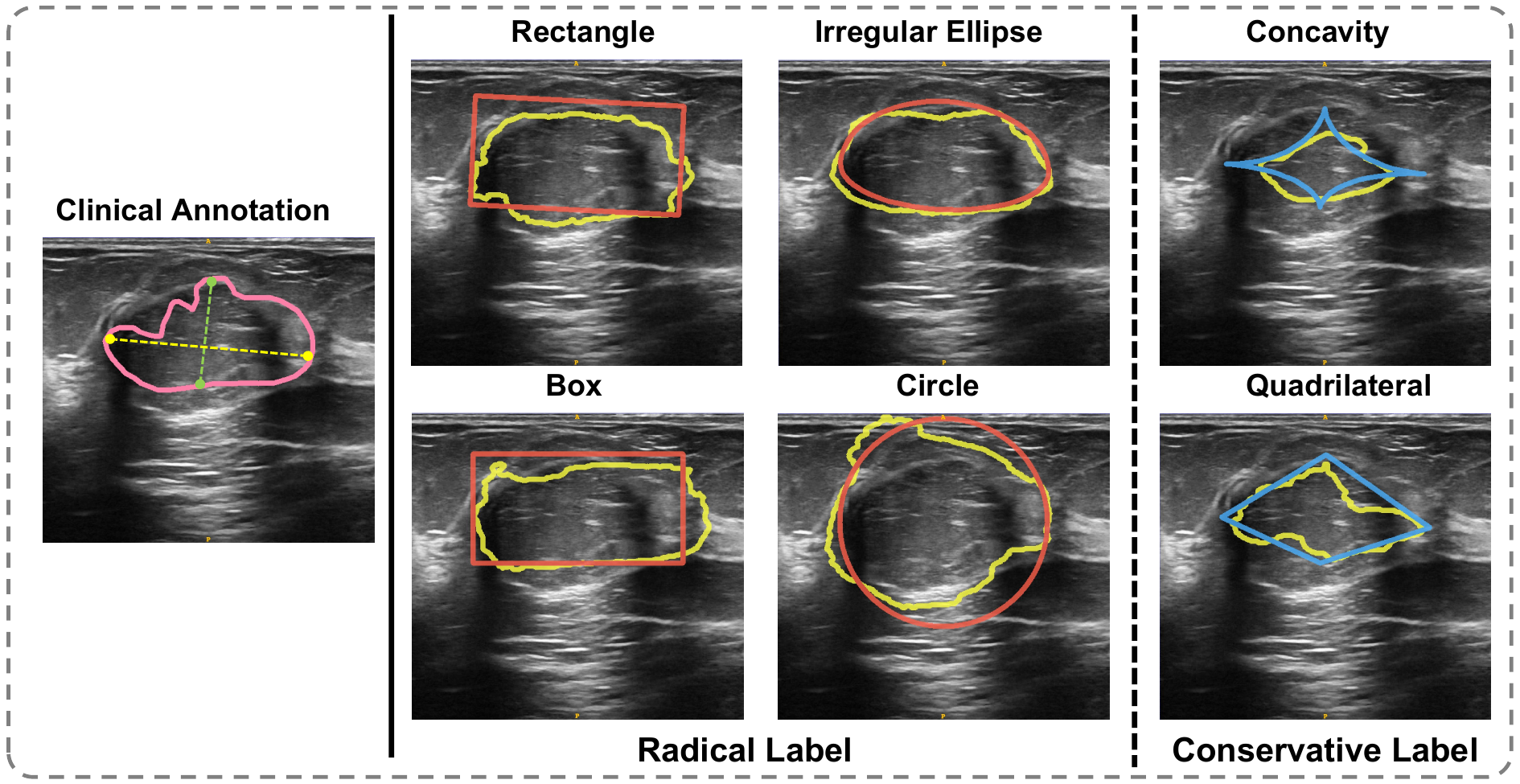}
    \caption{\small{Visualization of clinical aspect ratio annotation and the corresponding pseudo labels. The pink lines depict the edges of the ground truth, the orange lines represent the edges of the radical labels, the blue lines indicate the edges of the conservative labels, and the yellow lines display the predicted edges generated by the model trained using the respective pseudo label.}}
    \label{fig:intro}
\end{figure}

To alleviate the burdens associated with image annotation, endeavors have been devoted in exploring the use of readily available weak or sparse annotations, such as image-level annotations \cite{papandreou2015weakly, meng2019weakly}, scribbles \cite{lin2016scribblesup}, bounding boxes \cite{dai2015boxsup} and point annotations \cite{bearman2016s}, as alternatives for training networks. Nevertheless, even with their relatively easier acquisition compared to accurate annotations, collecting large-scale datasets with such annotations remains labor-intensive and time-consuming. Furthermore, these annotations may not sufficiently reflect the size and spatial information of the lesions, potentially misleading the model training process. 
These challenges are amplified in the context of ultrasound images due to their low resolution and complex lesion structures, which makes accurate annotation even more challenging. The significant variations in lesion shapes further complicate the use of existing sparse annotations for effective training. Consequently, there is an urgent need to explore more efficient strategies for sparse annotation and develop methods that can leverage these annotations for model training.

During ultrasound nodule diagnosis, doctors often measure the aspect ratio of nodules for clinical analysis. These aspect ratio annotations play an important role in providing information about the length, width, and location of the nodules and are widely available in  hospital picture archiving and communication systems (PACS). Based on this, we propose using aspect ratio annotations for training the model. Figure~\ref{fig:intro} provides a visual representation of the clinical aspect ratio annotation for nodules on ultrasound image. The longest line on the largest cross-section of the nodule is defined as the nodule's length, and the maximum distance perpendicular to this length is defined as the nodule's width \cite{moon2011taller}.

{\color{black}Sparse annotations cannot provide enough supervision signal for network training directly, necessitating the generation of pseudo-labels. Some researchers have explored using geometric shapes as pseudo-labels for model training. For instance, \textit{Zlocha et al.} \cite{zlocha2019improving} generated quadrilaterals by connecting adjacent endpoints of RECIST annotations, treating these as the foreground and considering the remaining area within the RECIST annotation’s bounding box as potential background. \textit{Tan et al.} \cite{tang2021weakly} created ellipses based on RECIST diameters for pseudo-labels, and \textit{Mahani et al.} \cite{mahani2022bounding} generated pseudo-labels by populating pixels within bounding box annotations. Despite these developments, there's a lack of extensive exploration into how geometric shapes correlate with lesion morphology and how this understanding could enhance training.}

In this study, we collected two clinically ultrasound datasets (thyroid and breast), accompanied by clinical annotations. Given the sparse nature of clinical annotations, which are not directly suitable for training, we further investigated the possibility of creating pseudo-labels. We generated basic geometric shapes including rectangle, irregular ellipse, quadrilateral, box, circle, and concavity based on these clinical annotations. We then filled these shapes to serve as pseudo labels for further training (see Figure \ref{fig:intro}).
Through analyzing the pseudo-labels generated directly based on clinical annotations, we have made two observations:

\textbf{Observation 1:}
``\emph{{These shapes can be divided into two distinct categories.}}'' Given that the lesions usually present convex shapes with irregular boundaries\cite{alexander2004thyroid}, one category (such as quadrilaterals) tends to underestimates lesions, leading to missed detections. We refer to this category as conservative labels. On the other hand, the second category of labels (such as circles) tends to overestimates lesions, resulting in false positives and including non-lesion areas. We term these as radical labels. This leads us to question: \textit{Can we develop a method that effectively combines radical and conservative labels to enhance their complementarity}? 

\textbf{Observation 2:}
``\emph{Using filled shapes as pseudo-labels may potentially mislead the model training process.}'' Directly using conservative labels or radical labels for model training without any additional processing can result in under-segmentation or over-segmentation in the predicted results. To achieve more accurate segmentation results, it is crucial to explore methods that address these issues and strike a balance between under-segmentation and over-segmentation (see Figure \ref{fig:intro}). This prompts us to ask: \textit{Can we find a novel approach to alleviate the issues of over-segmentation and under-segmentation}? 

Inspired by these observations, we introduce a novel asymmetric learning framework, called the \underline{\textbf{C}}onservative-\underline{\textbf{R}}adical-\underline{\textbf{B}}alance network (\textit{\textbf{CRBNet}}), which is based on clinical annotations. 
Our proposed model comprises two subnetworks, each trained using conservative labels and radical labels respectively. We first propose a conservative-radical-balance strategy (\textbf{CRBS}) to address the issues of over-segmentation and under-segmentation. This strategy applies conservative settings to models trained with conservative labels and radical settings to those trained with radical labels. Additionally, we introduce a module named Inconsistency-Aware Dynamically Mixed Pseudo Labels Supervision (\textbf{IDMPS}), which dynamically generates pseudo labels targeting specific regions where discrepancies exist between the radical and conservative labels. We also introduce a novel loss function, known as the clinical anatomy prior loss, which utilizes the spatial prior knowledge provided by the clinical annotations.

Extensive experiments are conducted on two clinically ultrasound datasets (thyroid and breast). The proposed framework achieves state-of-the-art performance in nodule segmentation compared to other relevant methods in clinical annotation settings. In summary, our contributions are as follows:

1) We develop an asymmetric weak-supervised segmentation framework based on aspect ratio annotations which can be directly obtained in ultrasound clinical examination. To the best of our knowledge, this is the first study to specifically explore the utilization of aspect ratio annotations from ultrasound clinical diagnoses for automated nodule segmentation.

2) We clearly identify and define two prevalent but distinct types of pseudo labels directly generated by clinical annotations. Our research presents interesting findings and conducts insightful discussions regarding these two types of pseudo labels. 

3) We have constructed two clinically ultrasound datasets (thyroid and breast), accompanied by clinical annotations. These datasets will be made publicly available upon acceptance of this paper for publication.

\section{Related Work}

\subsection{Medical Image Segmentation}
Medical image segmentation has evolved significantly over time, from traditional image processing techniques to deep learning-based methods \cite{lin2023dbganet,yun2023spectr}. In recent years, the fully convolutional network (FCN) \cite{long2015fully} based encoder-decoder architecture has emerged as the dominant paradigm for medical image segmentation with prominent examples including U-Net \cite{unet2015MICCAI} and V-Net \cite{milletari2016v}. The field continues to progress as researchers devise various methods and techniques to enhance segmentation performance. These methods and techniques can be grouped as follows: 1) The creation of innovative architectures and network components, such as Transformer networks, multi-scale networks and pyramid networks~\cite{hatamizadeh2022unetr,chen2021transunet,dalmaz2022resvit,zhou2019unet++,feng2020cpfnet,li2023erdunet}. 2) The development of novel data-driven techniques, including data augmentation methods and the use of pretrained networks\cite{xu2020automatic,chaitanya2021semi,iglovikov2018ternausnet}, which leverage large volumes of data to boost model performance and generalizability. 3) The development of model extension methods\cite{dolz2018hyperdense,yu2019uncertainty}, such as multi-modal image segmentation that integrates information from different imaging modalities, and uncertainty modeling, which estimates the reliability of segmentation predictions. 4) The use of advanced loss functions\cite{kervadec2019boundary,karimi2019reducing,yeung2022unified,luo2022semi}, which significantly enhances the accuracy of image segmentation. Despite these significant advancements, a prevailing challenge persists: most of these advancements require the availability of large volumes of finely annotated data. However, acquiring such large-scale, meticulously annotated medical data is a laborious and time-consuming endeavor. More importantly, this process necessitates the application of domain-specific knowledge, particularly from professionals in the field.

In recent years, deep convolutional neural networks have been widely used in ultrasound nodule diagnosis\cite{avola2021multimodal}. \textit{\color{black}Hu et al.} \cite{hu2019automatic} and \textit{\color{black}Ma et al.} \cite{ma2017ultrasound} were among the pioneers who employed convolutional neural networks for breast mass segmentation and thyroid nodule segmentation. To further enhance the segmentation performance of CNNs, researchers have explored modifications to the model architecture. \textit{\color{black}Xie et al.} \cite{xie2018breast} improved the Mask R-CNN\cite{he2017mask} to enhance its segmentation performance. \textit{\color{black}Byra et al.} \cite{byra2020breast} introduced attention mechanisms into the model architecture. Furthermore, various methods have also been explored to refine the segmentation process. For instance, \textit{\color{black}Ying et al.} \cite{ying2018thyroid} proposed a cascaded approach that involves training multiple CNN models iteratively. Each model utilizes the output of the previous model to progressively improve the accuracy of the segmentation results. \textit{\color{black}Karami et al.} \cite{karami2018adaptive} introduced an Adaptive Polar Active Contour algorithm, significantly improving IJV tracking by dynamically adjusting to frame-based segmentation results. Nevertheless, it is worth noting that these methods heavily rely on meticulously annotated labels generated by medical professionals, making the task both time-consuming and labor-intensive.

\subsection{Weakly-Supervised Medical Image Segmentation}
In the realm of image segmentation, weakly-supervised learning has been widely investigated to reduce the cost of annotation. Specifically, weakly supervised learning in medical image segmentation aims to learn segmentation using weak annotations, which can be classified into four types: image-level labels \cite{patel2022weakly}, bounding boxes \cite{du2023weakly}, points \cite{zhai2023pa}, and scribbles \cite{can2018learning,liu2022weakly}. 
{\color{black}For example, \textit{Pate et al.} \cite{patel2022weakly} introduced a novel learning strategy that enhances class activation maps (CAMs) using self-supervision in multi-modal image scenarios, thereby improving semantic segmentation under weakly supervised conditions. \textit{Du et al.} \cite{du2023weakly} proposed an innovative box-supervised segmentation framework for medical imaging that integrates geometric priors and contrastive similarity, specifically addressing the challenges posed by complex segment shapes and imaging artifacts. \textit{Zhai et al.} \cite{zhai2023pa} presented PA-Seg, a two-stage weakly supervised learning framework tailored for 3D medical image segmentation, which utilizes minimal point annotations and advanced regularization strategies to boost segmentation accuracy and effectively manage unannotated regions. \textit{Wang et al.} \cite{wang2023s} introduced a weakly-supervised polyp segmentation framework that employs scribble labels and merges spatial and spectral features with entropy-guided pseudo labels, aiming to enhance robustness and tackle the inherent challenges in medical imaging.}

Image-level labels provide a global understanding of the image content but lack precise details about object location and shape. While bounding boxes offer location and rough shape information, accurately delineating the object within those boundaries can be challenging. Points provide precise object localization but do not capture the object's extent and shape. Scribbles, like bounding boxes, offer some shape information but are not as comprehensive as full segmentation masks.

Recently, researchers have been exploring the use of RECIST (Response Evaluation Criteria in Solid Tumors) annotations for weakly supervised segmentation. However, since RECIST annotations cannot be directly used to train deep learning models, a crucial step in this approach is to generate pseudo-labels. Most previous works have relied on GrabCut~\cite{rother2004grabcut} to generate initial pseudo labels for fully supervised training.  GrabCut is an unsupervised image segmentation algorithm that utilizes graph theory and energy minimization. It requires a reliable trimap for each image, which contains information about the background (BG), foreground (FG), probably background (PBG), and probably foreground (PFG) information for each image. Several previous works\cite{cai2018accurate,wang2022recistsup} that were based on GrabCut have adopted a method of iteratively updating the trimap and pseudo label according to the model output. However, this approach has been found to be time-consuming during the training process. \textit{\color{black}Agarwal et al.} \cite{agarwal2020weakly} introduced a co-segmentation framework based on weak supervision, which can effectively capture shared semantic information from a pair of CT scans. However, they did not address the issue of handling the noisy pseudo labels generated by GrabCut. 
\textit{\color{black}Tang et al.} \cite{tang2021weakly} utilized ellipse masks as pseudo labels and refined the segmentation boundaries using the regional level set (RLS) loss. The approach introduced by \cite{zhou2023recist} consists of two models trained with a co-training strategy. This strategy leverages a consistency loss to contrast their predictions. However, simply using geometric shapes as pseudo labels can indeed lead to over-segmentation and under-segmentation issues. Additionally, employing a consistency strategy to compare the predictions of different subnets can potentially lead to error accumulation. \textit{\color{black}Zhou et al.} \cite{zhou2023recist2} introduced a weakly-supervised learning framework based on RECIST. Within this framework, they established a trimap that divides lesion slices into three distinct regions. By applying different processing strategies to these regions, they were able to enhance the segmentation performance. The methods mentioned above are primarily proposed for lesion detection in CT data. However, the lower resolution and the presence of irregular and tiny lesions in ultrasound images can degrade the performance of these models.

{\color{black}
\subsection{Learning Segmentation with Noisy Labels}
Numerous studies have demonstrated that label noise can have a substantial impact on network training, leading to a decrease in the accuracy of learned models \cite{zhang2018generalized, han2018co, zhang2020robust, northcutt2021confident, xu2022anti, zhu2019pick}. In response to this challenge, many efforts have been made to improve the robustness of model from noisy labels. Some researchers have focused on developing noise-robust loss functions to mitigate the impact of label noise on model performance. For instance, \textit{\color{black}Zhang et al.} \cite{zhang2018generalized} introduced a noise-robust loss function that builds upon mean absolute error (MAE) and categorical cross entropy (CCE), aiming to enhance deep neural network (DNN) training across large-scale datasets with noisy annotations. Several studies explored the use of pixel-wise loss correction to address this problem. \textit{\color{black}Han et al.} \cite{han2018co} proposed a co-teaching strategy where two networks are simultaneously trained, and each network selects reliable pixels with small loss values to supervise the training of the other network. \textit{\color{black}Zhang et al.} \cite{zhang2020robust} extended co-teaching \cite{han2018co} by training three networks simultaneously, where each pair of networks collaboratively selects reliable pixels to supervise the training of the third network. \textit{\color{black}Fang et al.} \cite{fang2023reliable} introduced a reliability-aware sample selection strategy that employs knowledge distillation to mitigate the impact of label noise on model training, thereby enhancing model performance. Inspired by confident learning in image level classification tasks \cite{northcutt2021confident}, \textit{\color{black}Xu et al.} \cite{xu2022anti} remold confident learning to generate a pixel-wise label error map and utilize it to refine the noisy label during network training. On the other hand, some methods focus on image-level noise estimation and learning. For example, \textit{\color{black}Zhu et al.} \cite{zhu2019pick} introduced an image-level label quality evaluation strategy that selectively picks high-quality samples and utilizes them to enhance the training process.
}
\begin{figure*}[htbp] 
	\centering
	\includegraphics[width=0.9\linewidth]{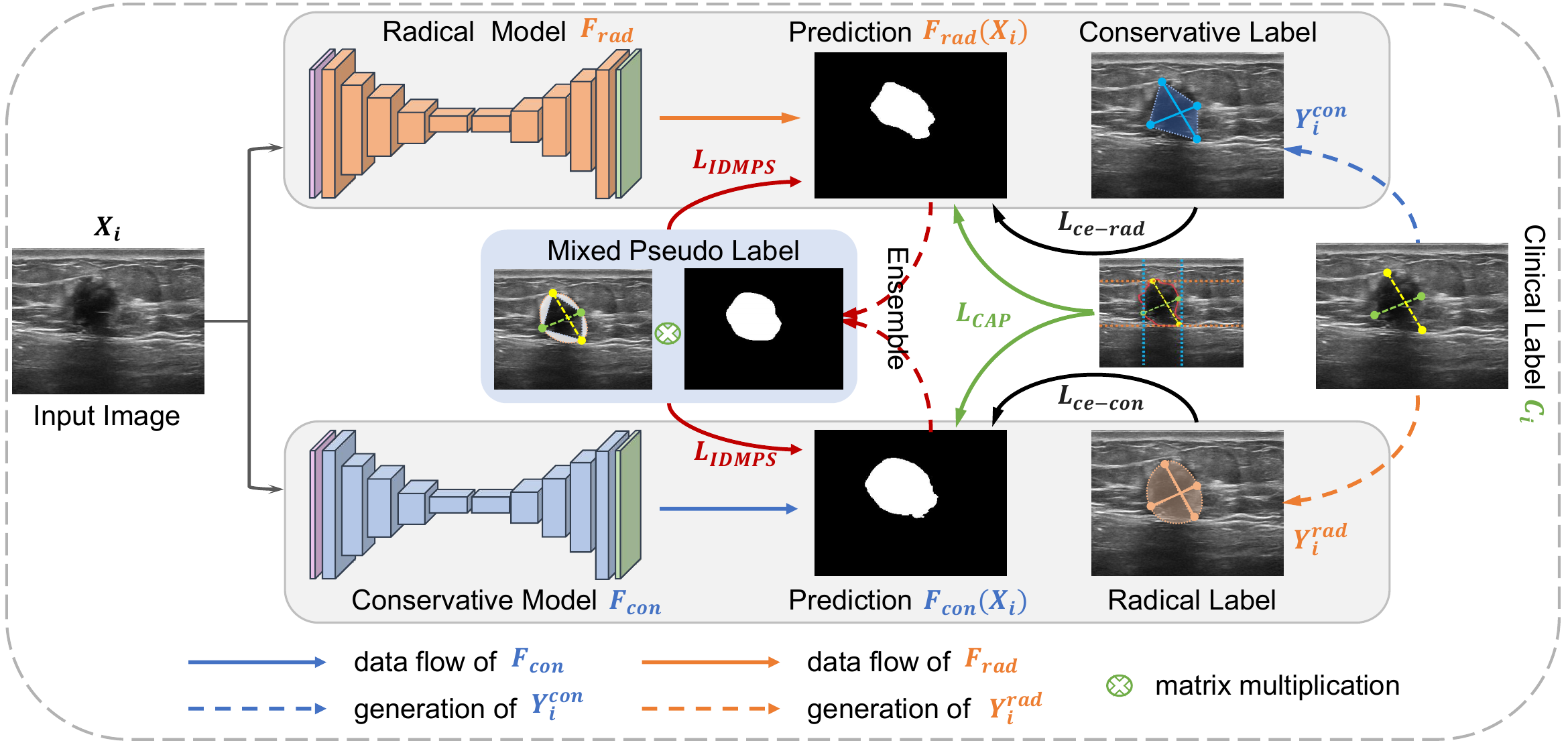}
	\caption{The overall framework of our proposed method. {\color{black} The areas highlighted with a gray background represent the CRB strategy, while those with a blue background correspond to the IDMPS module. The conservative model and conservative label are indicated in blue, whereas the radical model and radical label are depicted in orange.}}
	\label{fig:method} 
\end{figure*}

\section{Proposed Method}
\subsection{Overview}
The proposed asymmetric learning framework for weakly supervised medical image segmentation using clinical annotations is illustrated in Figure \ref{fig:method}. The framework consists of two main steps: mask generation and collaborative supervised learning framework. In the first step, we generate both conservative and radical labels based on the clinical annotation results for each image. In the second step, we introduce a collaborative supervised learning framework, which includes a Conservative Radical Balance Strategy (CRBS) and an Inconsistency-Aware Dynamically Mixed Pseudo Labels Supervision (IDMPS) module.
To leverage the location prior information of lesions provided by clinical annotations, we incorporate a clinical anatomy prior loss into the training process.

In the following sections, we present our method from two perspectives: mask generation and the collaborative supervised learning framework.

\textbf{Notation.} Following the literature, the primary aim of this investigation is to train a model using clinical annotations, with the ultimate goal of accurately classifying pixels as lesions and non-lesions in ultrasound images. Given $m$ images with clinical annotations as labels, we represent the images as $\mathbf{X}_{1}, \mathbf{X}_{2}, \ldots, \mathbf{X}_{m}\in\mathbb{R}^{H_{0}\times W_{0}}$, while their corresponding clinical annotations are denoted as $\mathbf{C}_{1},\mathbf{C}_{2},\ldots,\mathbf{C}_{m}\in\mathbb{B}^{H{0}\times W_{0}}$. To denote the generated radical and conservative labels, we use $\mathbf{Y}_{1}^{rad},\mathbf{Y}_{2}^{rad},\ldots,\mathbf{Y}_{m}^{rad}\in\mathbb{B}^{H{0}\times W_{0}}$ and $\mathbf{Y}_{1}^{con},\mathbf{Y}_{2}^{con},\ldots,\mathbf{Y}_{m}^{con}\in\mathbb{B}^{H{0}\times W_{0}}$, respectively. Note that the symbols $H_{0}$ and $W_{0}$ represent the height and width of the images correspondingly.
\subsection{Mask Generation}
In this work, we aim to generate basic geometric shapes by directly constructing them from clinical annotations, and then filling these shapes to serve as pseudo labels for further training. As shown in Figure \ref{fig:intro}, we generate six basic shapes based on clinical annotation. Each clinical annotation consists of a major axis and a minor axis that are perpendicular to each other. By connecting the endpoints of these axes and filling the enclosed area, we can generate a quadrilateral mask. Moreover, by identifying the smallest rectangular area that covers the foreground pixels within the clinical annotation, we obtain a bounding box that can be filled to create a rectangular mask. Similarly, a rotated rectangle mask can be obtained by filling the pixels inside a rotated bounding box. To generate a circular mask, we find the minimum enclosing circle that encompasses all the annotation points, and then fill the interior area of the circle. Additionally, an irregular elliptical mask can be generated by connecting curved lines between each pair of adjacent endpoints. Likewise, the concavity mask can be created by connecting curved lines in a reverse manner between each pair of adjacent endpoints. Notably, the concavity mask and quadrilateral mask are classified as conservative labels, while the box mask, rectangle mask, circular mask, and elliptical mask fall into the category of radical labels.
\subsection{Collaborative Supervised Learning Framework}
As illustrated in Figure \ref{fig:method}, our framework primarily consists of two subnetworks supervised by conservative labels and radical labels, respectively. We utilize the irregular elliptical masks as the radical labels and employ the quadrilateral masks as the conservative labels. Nonetheless, it has been observed that the model trained with radical labels may generate over-segmented outputs, whereas the one trained with conservative labels may produce under-segmented outputs. This suggests that finding a balance between these two types of labels is critical for achieving optimal segmentation performance. To alleviate this issue, we propose two strategies: the Conservative-Radical-Balance Strategy (CRBS) and the Inconsistency-Aware Dynamically Mixed Pseudo Labels Supervision (IDMPS). To leverage the spatial prior information provided by clinical annotations, we propose using the position of the target anatomical region indicated by clinical annotations to constrain the spatial location of the outputs generated by the two subnetworks. 
\subsubsection{CRBS}
In our study, we perform training of the object conservative and radical models using distinct relative misclassification costs. Specifically, we use positive relative misclassification costs for model supervised by radical labels and negative relative misclassification costs for model supervised by conservative labels to achieve the balance between false positives and false negatives during the training process. To illustrate, we use the widely adopted cross-entropy loss for the $i$-th image (where $1\leq i \leq m$): 
\begin{equation}
\begin{aligned}
{\color{black}
\mathcal{L}_{ce\text{-}con}(\mathbf{X}_i, \mathbf{Y}_i^{rad}; \mathcal{F}_{con}) = -\sum_{k=0}^{1}\sum_{z=1}^{H_0\times W_0} w^{con}_k  q_{z,k} \text{log} p_{z,k},}\\
{\color{black}
\mathcal{L}_{ce\text{-}rad}(\mathbf{X}_i, \mathbf{Y}_i^{con}; \mathcal{F}_{rad}) =  -\sum_{k=0}^{1}\sum_{z=1}^{H_0\times W_0} w^{rad}_k  q_{z,k} \text{log} p_{z,k}.}\\
\end{aligned}
\end{equation}
In this context, $\mathcal{L}_{ce\text{-}con}$ denotes the conservative cross-entropy loss, whereas $\mathcal{L}_{ce\text{-}rad}$ represents the radical cross-entropy loss. Moreover,  $\mathcal{F}_{con}$ and $\mathcal{F}_{rad}$ respectively denote the model supervised with radical and conservative labels. Additionally, the notation $p_{z,k}$ represents the probability of the $z$-th pixel belonging to the $k$-th class, whereas $q_{z,k}$ refers to the corresponding label of that pixel. Next, we define the weights for different classes, ${w}_{k}^{con}$ and ${w}_{k}^{rad}$, as follows:
\begin{equation}
w^{con}_k=
\begin{cases}
\alpha & k = 0\\
1 & k = 1
\end{cases} \quad \text{and} \quad
w^{rad}_k=
\begin{cases}
1 & k = 0\\
\alpha & k = 1.
\end{cases}
\end{equation}
By adjusting the hyperparameter $\alpha$, we are able to induce the model towards either conservative or radical predictions. The total loss $\mathcal{L}_{sup}$ of the proposed $CRB$ strategy is computed based on all labeled samples by summing up their respective losses. Mathematically, we can define $\mathcal{L}_{sup}$ as follows:
\begin{equation}
\begin{aligned}
\mathcal{L}_{sup}(\mathcal{F}_{rad},\mathcal{F}_{con}) = \sum_{i=1}^m \Bigg[\mathcal{L}_{ce-con}\left(\mathbf{X}_{i}, \mathbf{Y}_{i}^{rad}; \mathcal{F}_{con}\right) \\
+\mathcal{L}_{ce-rad}\left(\mathbf{X}_{i}, \mathbf{Y}_{i}^{con}; \mathcal{F}_{rad}\right)\Bigg].
\end{aligned}
\label{eqn:sup}
\end{equation}
\subsubsection{IDMPS}
IDMPS is to combine the outputs of the two subnetworks, which provides a trade-off between over-segmentation and under-segmentation. Given that the ground truth often falls somewhere between the conservative and radical classifications, we propose to dynamically mix the predictions of the conservative model and the radical model in the inconsistent region between the two types of labels, so as to generate hard pseudo labels. For the $i$-th image, the generated pseudo label $\mathbf{Y}_i^{PL}$ can be defined as:
\begin{equation}
\mathbf{Y}_i^{PL}=\operatorname{argmax}\left[\beta \times \mathcal{F}_{con}(\mathbf{X}_i)+(1.0-\beta) \times \mathcal{F}_{rad}(\mathbf{X}_i)\right],
\label{eqn:pl}
\end{equation}
where $\beta$ is a randomly generated value between 0 and 1, inclusive. Then, we use the generated pseudo labels to separately supervise ${F}_{con}$ and ${F}_{rad}$ for network training, only focusing on pixels where $\mathbf{Y}_{i}^{rad}$ and $\mathbf{Y}_{i}^{con}$ are inconsistent. The inconsistent region mask between the two types of labels, $\mathbf{M}_i$ can be defined as:
\begin{equation}
\mathbf{M}_{i}=\mathbf{Y}_{i}^{rad} \oplus \mathbf{Y}_{i}^{con} \in \mathbb{B}^{H_{0} \times W_{0}}.
\label{eqn:m}
\end{equation}
Therefore, the pseudo labels supervision is defined as:
\begin{equation}
\begin{aligned}
\mathcal{L}_{IDMPS}(\mathcal{F}_{rad},\mathcal{F}_{con}) = \sum_{i=1}^m \Bigg[\mathcal{L}_{ce-m}\left(\mathbf{X}_{i}, \mathbf{Y}_{i}^{PL}, \mathbf{M}_{i} ; \mathcal{F}_{rad}\right) \\
+\mathcal{L}_{ce-m}\left(\mathbf{X}_{i}, \mathbf{Y}_{i}^{PL}, \mathbf{M}_{i} ; \mathcal{F}_{con}\right)\Bigg],
\end{aligned}
\label{eqn:pls_all}
\end{equation}
where $\mathcal{L}_{ce-m}$ is defined as:
\begin{equation}
\mathcal{L}_{ce-m}\left(\mathbf{X}_{i}, \mathbf{Y}_{i}^{PL}, \mathbf{M}_{i} ; \mathcal{F}\right)=-\sum_{k=1}^{2} \sum_{z=1}^{H_{0} \times W_{0}} \mathbf{M}_{j, z} q_{z, k} \log p_{z, k}.
\end{equation}
This strategy achieve a balance between over-segmentation and under-segmentation. Moreover, the use of the random blending weight $\beta$ during each iteration enhances the diversity of pseudo labels and alleviates the issue of one model misleading another. 
Different from consistency learning, there are no explicit constraints applied to ensure the predictions of both subnetworks become similar to each other.
\begin{algorithm}[t]
\caption{The core learning algorithm of our method.}
\label{alg:algorithm}
\hspace*{0.02in}{\textbf{{Input}}}: image $\mathbf{X}_1$, $\mathbf{X}_2$, ..., $\mathbf{X}_m$ and their clinical annotation $\mathbf{C}_1$, $\mathbf{C}_2$, ..., $\mathbf{C}_m$\\
\hspace*{0.02in}{\textbf{{Output}}}: radical model's parameter $\theta_{1}$ and conservative model's parameter $\theta_{2}$
\begin{algorithmic}[1]
\STATE $\mathcal{F}_{rad}\left(x\right)$ = radical model with parameter $\theta_{1}$ \\
\STATE $\mathcal{F}_{con}\left(x\right)$ = conservative model with parameter $\theta_{2}$ \\
\WHILE{stopping criterion not met:}
\STATE Sample batch $b = (\mathbf{X}_{i},\mathbf{C}_{i})$, $i \in \{1, ..., N\}$ where $N$ denotes the batch size
\STATE Generating radical label $\mathbf{Y}_{i}^{rad}$ and conservative label $\mathbf{Y}_{i}^{con}$ based on $\mathbf{C}_{i} \in b$ \\
\STATE Computing the inconsistent region mask $\mathbf{M}_{i}$ between radical label $\mathbf{Y}_{i}^{rad}$ and conservative label $\mathbf{Y}_{i}^{con}$ according to Eqn. (\ref{eqn:m}) \\
\STATE Computing predictions of conservative model $\mathcal{F}_{con}(\mathbf{X}_i)$ and predictions of radical model $\mathcal{F}_{rad}(\mathbf{X}_i)$, where $i \in \{1, ..., N\}$
\STATE Generating dynamically mixed pseudo labels
$\mathbf{Y}_{i}^{PL}$ based on $\mathcal{F}_{con}(\mathbf{X}_i)$ and $\mathcal{F}_{rad}(\mathbf{X}_i)$ using Eqn. (\ref{eqn:pl})\\
\STATE Calculating supervision loss $\mathcal{L}_{sup}$ using Eqn. (\ref{eqn:sup}) \\
\STATE Calculating pseudo labels supervision loss $\mathcal{L}_{IDMPS}$ using Eqn. (\ref{eqn:pls_all}) \\
\STATE Calculating Clinical Anatomy Prior loss $\mathcal{L}_{CAP\_tot}$ using Eqn. (\ref{eqn:cap}) \\
\STATE $\mathcal{L}_{{total}}=\mathcal{L}_{{sup}}+\lambda_{1}\mathcal{L}_{IDMPS}+\lambda_{2}\mathcal{L}_{CAP_{tot}}$
\STATE Computing gradient of loss function $\mathcal{L}_{total}$ and update network parameters $\theta_{1}$, $\theta_{2}$ by back propagation. \\
\ENDWHILE \\
\RETURN  $\theta_{1}$, $\theta_{2}$
\end{algorithmic}
\label{alg:algsummary}
\end{algorithm}
\subsubsection{Clinical Anatomy Prior Loss}
{For the $i$-th image, $\mathcal{F}(\mathbf{X}_i) \in(0,1)^{2 \times H_{0} \times W_{0}}$} denotes the predicted probability map generated by the network. Next, the spatial location of $\mathcal{F}(\mathbf{X}_i)$ are defined as follows:
\begin{equation}
\begin{aligned}
\text{pre\_posi}_x(\mathbf{X}_i; \mathcal{F}) &= \max_{k=1}^2\max_{h=1}^{H_0} \mathcal{F}(\mathbf{X}_i)_{k,h,:} \in {(0,1)}^{H_{0}},\\
\text{pre\_posi}_y(\mathbf{X}_i; \mathcal{F}) &= \max_{k=1}^2\max_{w=1}^{W_0} \mathcal{F}(\mathbf{X}_i)_{k,:,w} \in {(0,1)}^{W_{0}},\\
\end{aligned}
\end{equation}
where $\text{pre\_posi}_x$ and $\text{pre\_posi}_y$ represent the maximum probabilities along the x-axis and y-axis, respectively, capturing the spatial information. Similarly, the position of the target anatomical region indicated by clinical annotation are defined as follows:
\begin{equation}
\begin{aligned}
\text{C\_posi}_x(\mathbf{C_i}) &= \max_{h=1}^{H_0} \mathbf{C_i}_{h,:} \in \mathbb{B}^{H_{0}},\\
\text{C\_posi}_y(\mathbf{C_i}) &= \max_{w=1}^{W_0} \mathbf{C_i}_{:,w} \in \mathbb{B}^{W_{0}}, 
\end{aligned}
\end{equation}
where $\text{C\_posi}_x$ and $\text{C\_posi}_y$ represent the position information of the target anatomical region provided by clinical annotations along the x-axis and y-axis, respectively, providing anatomical priors. We define the supervised loss for Clinical Anatomy Prior as commonly used dice loss:
\begin{equation}
\mathcal{L}_{\text{Dice}}\left(\mathbf{\hat{Y}}_{i}, \mathbf{Y}_{i}\right)=-\frac{1}{2}\sum_{k=0}^{1}\left(1-\frac{2 \sum_{z=1}^{H_{0} \times W_{0}} p_{z, k} q_{z, k} }{\sum_{z=1}^{H_{0} \times W_{0}} p_{z, k}+q_{z, k}}\right).
\end{equation}
Therefore, the Clinical Anatomy Prior loss of $i-$th image can be defined as:
\begin{equation}
\begin{aligned}
\mathcal{L}_{CAP}(\mathbf{X}_{i},\mathbf{C}_{i};\mathcal{F}) & = \mathcal{L}_{Dice}\left(\text{pre\_posi}_x(\mathbf{X}_i; \mathcal{F}), \text{C\_posi}_x(\mathbf{C_i})\right) \\
&+\mathcal{L}_{Dice}\left(\text{pre\_posi}_y(\mathbf{X}_i; \mathcal{F}), \text{C\_posi}_y(\mathbf{C_i})\right). \\
\end{aligned}
\end{equation}
Given that the output anatomical regions generated by the two subnetworks should be consistent along the x-axis and y-axis, we propose using the position of the target anatomical region indicated by clinical annotations onto the x-axis and y-axis to supervise the output anatomical regions of the two subnetworks, respectively. Therefore, the total Clinical Anatomy Prior loss is calculated on all the labeled samples, which can be defined as:
\begin{equation}
\begin{aligned}
\mathcal{L}_{CAP\_{tot}}(\mathcal{F}_{con},\mathcal{F}_{rad}) = \sum_{i=1}^m & \Bigg[\mathcal{L}_{CAP}(\mathbf{X}_i, \mathbf{C}_i; \mathcal{F}_{con}) \\
&+\mathcal{L}_{CAP}(\mathbf{X}_i, \mathbf{C}_i; \mathcal{F}_{rad})\Bigg]. \\
\end{aligned}
\label{eqn:cap}
\end{equation}
\begin{table*}[t!]
\small
\centering
\caption{Performance comparison with other methods. We evaluate the performance using four metrics: Dice Score, HD95, Jaccard, and ASD. The symbol * indicates a p-value of less than 0.05 (paired t-test) when comparing with the second-place method. The \textcolor{red}{*} symbol indicates a p-value of less than 0.05 (paired t-test) when comparing with the fully supervised setting.
}
\renewcommand\arraystretch{1.4}
\newcolumntype{P}[1]{>{\centering\arraybackslash}p{#1}}
\resizebox{1.\linewidth}{!}{
    \begin{tabular}{P{4.3cm}P{1.3cm}P{1.3cm}P{1.3cm}P{1.3cm}cP{1.3cm}P{1.3cm}P{1.3cm}P{1.3cm}}
        \specialrule{.8pt}{0pt}{2pt}
        \multirow{2}{*}{Method} &
        \multicolumn{4}{c}{Thyroid Ultrasound}  && \multicolumn{4}{c}{Breast Ultrasound} \\ 
        \cline{2-5} \cline{7-10} 
        &  DSC(\%)&HD95(pixel)&ASD(pixel)&Jaccard(\%)&&DSC(\%)&HD95(pixel)&ASD(pixel)&Jaccard(\%) \\
          \specialrule{.4pt}{2pt}{0pt}
          \multicolumn{10}{l}{\textit{Noisy label learning methods:}} \\
          CoTeaching~\cite{han2018co} & 71.2$\pm$24.0 & 38.7$\pm$72.5 &14$\pm$28.9 & 59.5$\pm$22.7 && 72.7$\pm$21.8 & 53.4$\pm$37.3 & 19.6$\pm$17.8 & 60.9$\pm$22.0 \\
          TriNet~\cite{zhang2020robust} & 70.8$\pm$23.9 & 35.9$\pm$72.4 &11.9$\pm$26.6 & 59$\pm$22.8 && 69.5$\pm$22.6 & 58.2$\pm$38.8 & 20.9$\pm$17.3 & 57.1$\pm$22.5 \\
          MTCL~\cite{xu2022anti} & 70.2$\pm$25.0 & 40.1$\pm$76.3 & 14$\pm$29.9 & 58.5$\pm$23.7 && 70.5$\pm$22.1 & 60.1$\pm$38.6 & 21$\pm$17.1 & 58.3$\pm$22.5 \\
          \textcolor{black}{GCE~\cite{zhang2018generalized}} & \textcolor{black}{72.6$\pm$23.1} & \textcolor{black}{42.6$\pm$82.3} & \textcolor{black}{14.5$\pm$35.3} & \textcolor{black}{61.0$\pm$22.6} && \textcolor{black}{71.1$\pm$22.1} & \textcolor{black}{61.2$\pm$44.8} & \textcolor{black}{21.8$\pm$20.2} & \textcolor{black}{58.9$\pm$22.4} \\
          \textcolor{black}{RMD~\cite{fang2023reliable}} & \textcolor{black}{69.5$\pm$25.2} & \textcolor{black}{46.6$\pm$80.9} & \textcolor{black}{15.7$\pm$28.8} & \textcolor{black}{57.8$\pm$24.0} && \textcolor{black}{73.9$\pm$22.2} & \textcolor{black}{52.8$\pm$38.2} & \textcolor{black}{18.7$\pm$17.9} & \textcolor{black}{62.6$\pm$22.8} \\
          \hdashline
          \multicolumn{10}{l}{\textit{Weakly supervised methods:}} \\
          EM~\cite{grandvalet2004semi} & 71.7$\pm$23.6 & 35.6$\pm$68.4 & 11.8$\pm$29.3 & 60$\pm$22.8 && 72.4$\pm$21.0 & 62.5$\pm$43.3 & 22.0$\pm$19.8 & 60.3$\pm$21.7 \\
          TV~\cite{javanmardi2016unsupervised} & 70.7$\pm$23.5 & 53.4$\pm$95.8 & 21.7$\pm$46.6 & 58.7$\pm$22.5 && 73.4$\pm$21.0 & 59.3$\pm$43.0 & 20.2$\pm$18.4 & 61.5$\pm$21.9 \\
          Mumford-Shah~\cite{kim2019mumford}  & 71.7$\pm$22.6 & 36.3$\pm$71.4 & 13.3$\pm$33.0 & 59.6$\pm$21.7 && 72.3$\pm$24.1 & 57.7$\pm$45.0 & 20.6$\pm$20.4 & 61.1$\pm$24.2 \\
          GatedCRF~\cite{obukhov2019gated}  & 68.6$\pm$25.2 & 39.1$\pm$74.6 & 14.4$\pm$33.4 & 56.7$\pm$23.7 && 73.2$\pm$20.3 & 57.7$\pm$39.6 & 21.1$\pm$21.3 & 61.1$\pm$21.5 \\
          WSSS~\cite{cai2018accurate}  & 73.3$\pm$21.3 & 37.2$\pm$67.1 & 14.1$\pm$27.9 & 61.4$\pm$21.7 && 74.7$\pm$19.5 & 59.8$\pm$41.9 & 22.2$\pm$20.4 & 62.8$\pm$20.8 \\
          RECISTSup~\cite{wang2022recistsup}  & 72.5$\pm$21.9 & 37.5$\pm$72.9 & 13.3$\pm$27.7 & 60.6$\pm$22.0 && 72.5$\pm$20.8 & 61.6$\pm$41.1 & 22.3$\pm$18.4 & 60.3$\pm$21.7 \\
          CoTraining~\cite{zhou2023recist}  & 73.4$\pm$23.7 & 39.2$\pm$74.0 & 15$\pm$31.4 & 62.2$\pm$23.2 && 68.3$\pm$22.0 & 58.9$\pm$37.5 & 21.8$\pm$17.7 & 55.4$\pm$21.3 \\
          \textcolor{black}{S$^2$ME~\cite{wang2023s}} & \textcolor{black}{70.4$\pm$24.7} & \textcolor{black}{39.7$\pm$76.4} & \textcolor{black}{12.9$\pm$28.7} & \textcolor{black}{58.7$\pm$23.3} && \textcolor{black}{72.4$\pm$20.5} & \textcolor{black}{55.0$\pm$36.0} & \textcolor{black}{19.6$\pm$17.8} & \textcolor{black}{60.1$\pm$22.4} \\
          \textcolor{black}{WSDAC~\cite{li2023weakly}} & \textcolor{black}{71.4$\pm$23.2} & \textcolor{black}{44.9$\pm$76.7} & \textcolor{black}{14.8$\pm$28.6} & \textcolor{black}{59.4$\pm$22.5} && \textcolor{black}{72.3$\pm$19.9} & \textcolor{black}{56.1$\pm$37.1} & \textcolor{black}{20.9$\pm$18.8} & \textcolor{black}{59.8$\pm$20.6} \\
          Ours & \textbf{76.5$\pm$22.5}* &\textbf{28.9$\pm$64.7}* & \textbf{8.7$\pm$22.1}* & \textbf{66.1$\pm$22.7}* && \textbf{76.6$\pm$\textbf{19.5}}*& \textbf{50.6$\pm$34.3}* & \textbf{18.8$\pm$18.0}* & \textbf{65.3$\pm$20.5}* \\ \hline 
          GrabCut\cite{rother2004grabcut} & 68.6$\pm$25.3 & 48.4$\pm$85.2 & 18.2$\pm$42.0 & 56.7$\pm$24.0 && 71$\pm$23.0 & 59.2$\pm$42.7 & 20.9$\pm$19.0 & 59.1$\pm$23.2 \\
          Fully Supervised & 74.7$\pm$23.8\textcolor{red}{*} & 35.3$\pm$75.3 & 13$\pm$31.5\textcolor{red}{*} & 64$\pm$23.7\textcolor{red}{*} && 76.4$\pm$20.6 & 53.2$\pm$43.2 & 18.8$\pm$19.6 & 65.5$\pm$22.4 \\
        \specialrule{.8pt}{0pt}{2pt}
    \end{tabular}
}
\label{tab:sota}
\end{table*}
Finally, the total objective function $\mathcal{L}_{total}$ can be summarized as:
\begin{equation}
\mathcal{L}_{{total}}=\mathcal{L}_{{sup}}+\lambda_{1}\mathcal{L}_{IDMPS}+\lambda_{2}\mathcal{L}_{CAP\_tot},
\label{tab:tol}
\end{equation}
where $\lambda_{1}$ and $\lambda_{2}$ are the weight factors for balancing these three terms. The training algorithm used in our proposed method is presented in Algorithm \ref{alg:algorithm}.

\section{Datasets And Experiments}
\subsection{Ultrasound Datasets}
The ultrasound dataset proposed in this study is composed of
two subsets: the Thyroid Ultrasound Dataset and the Breast Ultrasound Dataset. Throughout the data collection process,
we meticulously filtered out low-quality images and removed all irrelevant information, such as device details in the image boundary regions, through precise cropping. After cropping, the size of the thyroid ultrasound images is [817, 577], and the size of the breast ultrasound images is [575, 530]. Ultimately, we carefully selected a total of 844 images featuring thyroid nodules and 755 images featuring breast nodules for our experimental data. The Thyroid Ultrasound Dataset includes 422 left thyroid ultrasound images and 422 right thyroid ultrasound images. The Breast Ultrasound Dataset contains 604 left breast ultrasound images and 151 right breast ultrasound images. All thyroid and breast nodules in the dataset, along with their corresponding aspect ratio annotations, were manually segmented by two radiologists, each with over ten years of experience. Subsequently, the annotation results underwent thorough examination by a third radiologist with over twenty years of experience.

The final dataset was divided into training and testing sets. We used the training set to train the model and evaluated its performance on an independent testing set. We randomly divided the dataset into training and test sets, with a 4:1 ratio. The Thyroid Ultrasound Dataset consists of 675 images in the training
set and 169 images in the testing set. The Breast Ultrasound Dataset, on the other hand, includes 604 training images and 151 testing images. During the training phase, we ensured that each image in the training set was accompanied by its respective clinical annotation. Table \ref{tab:data} shows the descriptive statistics for the aforementioned datasets.
\begin{table}[htbp]
\footnotesize
\centering
\caption{Descriptive Statistics Of The Proposed Datasets.}
\label{one}
\renewcommand\arraystretch{1}
\begin{tabular}{c|cccccc}
\toprule
Datasets &Type & Left & Right & Train & Test & Total \\
\midrule
\specialrule{0pt}{1pt}{1pt}
Thyroid &nodule & 422  & 422   & 675        & 169        & 844         \\
Breast &nodule & 604  & 151   & 604        & 151        & 755         \\
\bottomrule
\end{tabular}
\label{tab:data}
\end{table}

\subsection{Experimental Setup}
\subsubsection{Implementation Details}
We implemented and executed our proposed method, along with other comparison methods, by PyTorch\cite{paszke2019pytorch} on an Ubuntu desktop equipped with an NVIDIA GeForce RTX 3090 GPU. For our backbone network, we adopted UNet \cite{unet2015MICCAI}, a widely recognized architecture in medical image segmentation. To ensure uniformity, all images were cropped to a size of 256 × 256 pixels and their intensity was rescaled to a range of 0-1 before feeding them into the network. To enhance the diversity of the training set, we employed random rotation and flipping techniques. For optimization, we utilized the SGD optimizer with weight decay of 0.0001 and momentum of 0.9. The batch size was set to 24, and the total number of iterations was 30,000. The initial learning rate was set at 0.01, and we applied the poly learning rate strategy \cite{luo2021efficient} for dynamic learning rate adjustment. In Eq.~\ref{tab:tol}, $\lambda_{1}$ is determined using a Gaussian ramp-up function and $\lambda_{2} = 0.3$. The determination of these weights was based on experimentation.
\subsubsection{Evaluation Criteria}
\begin{figure*}[htbp] 
	\centering
	\includegraphics[width=1.0\linewidth]{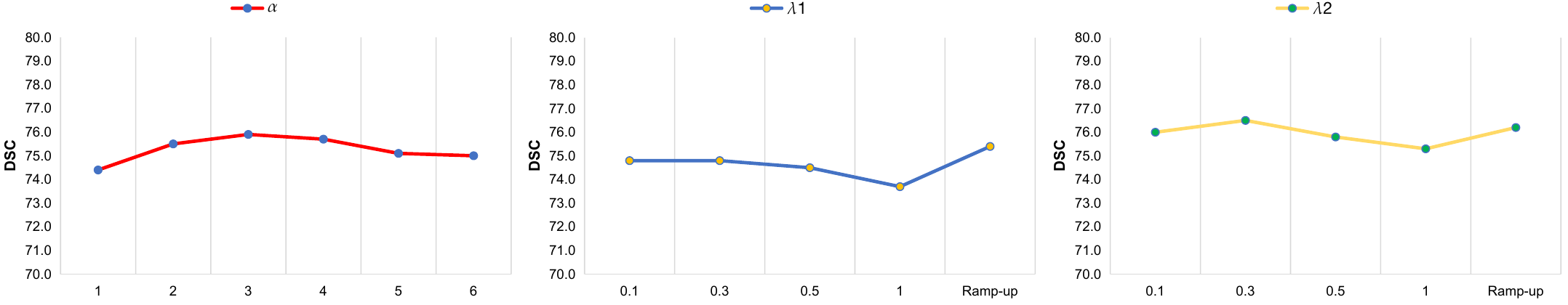}
	\caption{Performance sensitivity to hyper-parameter $\alpha$ and loss weights, $\lambda_{1}$ and $\lambda_{2}$ on the thyroid ultrasound dataset.}
	\label{fig:para} 
\end{figure*}
To quantitatively evaluate the performance of various segmentation methods, we employed several metrics in this experiment, including the Dice Similarity Coefficient (DSC), Jaccard Index, Average Surface Distance (ASD), and Hausdorff Distance at 95\% (HD95). DSC quantifies the level of agreement between the predicted segmentation and the ground truth, serving as an indicator of their similarity.. 
It is formulated as:
\begin{equation}
\operatorname{DSC}(A, B)=\frac{2|A \cap B|}{|A|+|B|},
\end{equation}
where $A$ represents the set of pixels in the predicted segmentation and $B$ represents the set of pixels in the ground truth. 
Similarly, the Jaccard Index is computed as the ratio of the intersection to the union of the predicted segmentation and the ground truth, assessing how accurately the predicted region aligns with the actual region. 
It is formulated as:
\begin{equation}
\operatorname{Jaccard}(A, B) = \frac{|A \cap B|}{|A \cup B|.}
\end{equation}
ASD measures the mean spatial distance between the predicted segmentation boundary and the ground truth boundary, providing an indication of the precision of the predicted boundaries. 
It is formulated as:
\begin{equation}
\operatorname{ASD}(A, B) = \frac{1}{|A|} \sum_{a \in A} \min_{b \in B} \lVert a - b \rVert.
\end{equation}
HD95 measures the maximum distance between the predicted segmentation and the ground truth segmentation. Unlike ASD, HD95 focuses on the worst-case scenario of boundary matching. 
The calculation formula for HD is as follows:
\begin{equation}
\operatorname{HD}(A, B) = \max(\operatorname{h}(A, B), \operatorname{h}(B, A))
\end{equation}
where,
\begin{equation}
\operatorname{h}(A, B) = \max(\sup_{a \in A} \inf_{b \in B} \lVert a-b \rVert, 0)
\end{equation}
represents the directed Hausdorff distance from set A to set B. Similarly, $\operatorname{h}(B, A)$ represents the directed Hausdorff distance from set B to set A. The HD95 is a modified version of the Hausdorff Distance that focuses on the 95$th$ percentile of distances between $A$ and $B$.
\begin{figure*}[htbp] 
	\centering
	\includegraphics[width=1.0\linewidth]{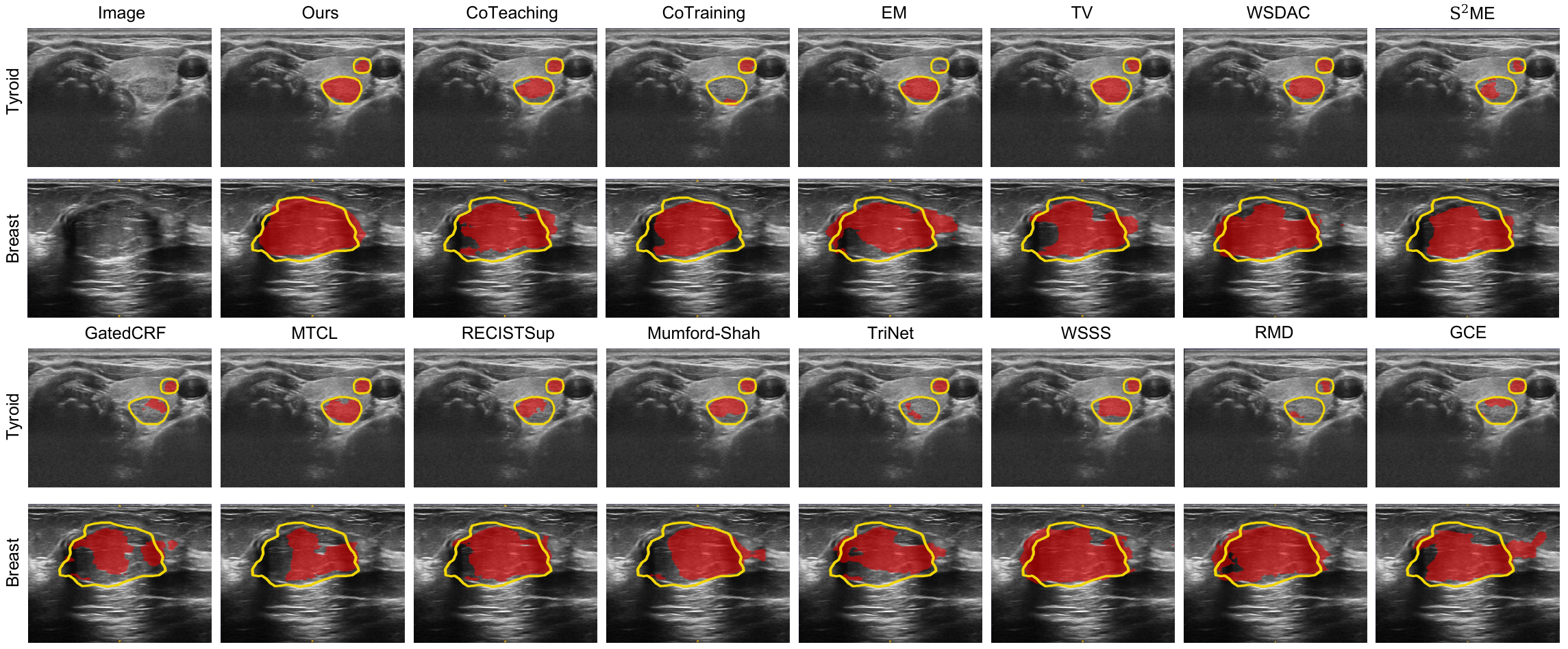}
	\caption{The visual segmentation examples of different comparison methods and the ground truth. The yellow curves show ground truth segmentation.}
	\label{fig:vis} 
\end{figure*}
\subsection{Annotation Cost Analysis}
\label{aca}
To evaluate the efficiency of aspect ratio annotations, we conducted a comparison study focusing on the cost of annotation. In this study, a seasoned doctor with over twenty years of experience performed two types of annotations on ten ultrasound images: accurate annotations and aspect ratio annotations. The average time taken for each annotation type was calculated to estimate the annotation cost. For thyroid ultrasound images, it took an average of 40 seconds for a accurate annotation, compared to just 5 seconds for an aspect ratio annotation. Similarly, for breast ultrasound images, the accurate annotation averaged 60 seconds per image, while the aspect ratio annotation required only 5 seconds per image. Experimental results indicate that annotating the aspect ratio of nodules in ultrasound images requires only 8\% to 13\% of the annotation cost needed for accurate annotations. Additionally, the advantages of aspect ratio annotation extend beyond just reduced annotation costs. Accurate annotations require experienced doctors to spend a considerable amount of time, while aspect ratio annotation of nodules is a routine procedure in clinical examinations. As a result, aspect ratio annotations are widely accessible in hospital picture archiving and communication systems (PACS), without adding extra workload for the doctors.

\subsection{Comparison With State-of-the-Arts}
\label{cwsota}
We evaluated the segmentation performance of our proposed framework and compared it with several state-of-the-art weakly supervised approaches: 1) GrabCut only (lower bound), 2) Entropy Minimization (EM)~\cite{grandvalet2004semi}, which reduces uncertainty in unannotated data to drive models towards making high-confidence predictions, 3) Total Variation (TV) loss~\cite{javanmardi2016unsupervised}, which promotes spatial smoothness by minimizing pixel variation among neighboring regions, 4) Mumford-Shah loss~\cite{kim2019mumford}, which balances segmentation smoothness and data fidelity to encourage segmentation continuity, 5) GatedCRF Loss~\cite{obukhov2019gated}, integrating pixel-level features and pairwise spatial relations using a Conditional Random Field (CRF) gated by model predictions for enhanced region coherence, 6) WSSS~\cite{cai2018accurate}, which iteratively updates pseudo-labels based on model output, 7) RECISTSup~\cite{wang2022recistsup}, which iteratively updates pseudo-labels during training, 8) CoTraining~\cite{zhou2023recist}, which trains two networks with distinct labels and imposes consistency constraints on their outputs{\color{black}, 9) S$^2$ME\cite{wang2023s}, which introduces a entropy-guided weakly-supervised polyp segmentation framework, 10) WSDAC\cite{li2023weakly}, which proposes a novel weakly supervised deep active contour model for nodule segmentation.} Additionally, as the pseudo-labels generated from clinical annotations may contain noise, we evaluated our method alongside other popular approaches for learning from noisy annotations: 1) CoTeaching\cite{han2018co}, where two networks exchange learned experiences and selectively teach each other using correctly learned instances, 2) TriNet\cite{zhang2020robust}, which trains three networks collaboratively to select reliable pixels that supervise the third network's training, 3) MTCL\cite{xu2022anti}, which utilizes confident learning to refine noisy labels{\color{black}, 3) GCE\cite{zhang2018generalized}, which proposes a noise robust loss function, 4) RMD\cite{fang2023reliable}, which utilizes  knowledge distillation to mitigate the impact of label
noise.} To ensure fair comparisons, we employed the widely adopted U-Net as the common backbone network for all methods in our experiments. {\color{black}It is important to note that all compared methods utilized aspect ratio annotations as labels. For methods that proposed their own pseudo-label generation techniques, we followed their specified approaches. For those without a specific pseudo-label generation method, we employed the commonly used GrabCut\cite{rother2004grabcut} to generate pseudo-labels for training.}

The quantitative evaluation results of these methods are summarized in Table \ref{tab:sota}. The results illustrate that our method outperforms other weakly supervised methods and noisy label learning methods, as depicted in Table \ref{tab:sota}. Our method achieved a promising Dice score of 0.765 on the Thyroid Ultrasound dataset, surpassing the results obtained by the fully supervised setting. Additionally, on the Breast Ultrasound dataset, our method achieved a Dice score of 0.766, which was slightly better than the Dice score obtained by the fully supervised setting. Figure \ref{fig:vis} shows some qualitative evaluation results. As illustrated in the figure, our proposed method achieves good segmentation performance.
\subsection{Ablation Experiments}
\subsubsection{Comparative Analysis of Different Components}
To assess the effectiveness of every proposed component, we carried out a sequence of experiments on the Thyroid Ultrasound dataset. The experimental results are presented in Table \ref{tab:components}. As shown in Table \ref{tab:components}, with the gradual introduction of components IDMPS, CRBS, and clinical anatomy prior loss, the performance consistently improves. The method achieved a Dice score of 0.677 when trained with conservative labels but exhibited a higher score of 0.731 when trained with radical labels. Additionally, when solely incorporating the IDMPS or $\mathcal{L}_{CAP}$, the Dice score improved to 0.754 and 0.744, respectively. Notably, the introduction of the CRB strategy on top of the IDMPS approach further boosted the Dice score to 0.759. Ultimately, the combination of IDMPS, CRB, and $\mathcal{L}_{CAP}$ yielded the most optimal performance, achieving an impressive Dice score of 0.765.
\begin{table}[t!]
\small
\centering
\caption{\small{results of the component variations. we measure the corresponding DSC, Hd95, ASD and Jaccard on the Thyroid Ultrasound dataset. $\dagger$: we trained the model solely using the conservative label. $\ddagger$: we trained the model solely using the radical label.}}
\renewcommand\arraystretch{1.4}
\newcolumntype{P}[1]{>{\centering\arraybackslash}p{#1}}
\resizebox{\linewidth}{!}{
	\begin{tabular}{cccccccc}
\specialrule{.8pt}{0pt}{2pt}
\multirow{2}{*}{Baseline} &
\multirow{2}{*}{IDMPS} &
\multirow{2}{*}{CRBS} &
\multirow{2}{*}{$\mathcal{L}_{CAP}$} &
\multicolumn{4}{c}{Evaluation Criteria} \\
\cline{5-8} 
& & & & $\uparrow$DSC(\%)&$\downarrow$HD95&$\downarrow$ASD&$\uparrow$Jaccard(\%) \\
\specialrule{.4pt}{2pt}{0pt}
\checkmark$^{\dagger}$ & & & & 67.7$\pm$27.2 & 34.1$\pm$69.1 & 14.1$\pm$37.4 & 56.3$\pm$25.2 \\
\checkmark$^{\ddagger}$ & & & & 73.1$\pm$24.3 & 47.2$\pm$80.3 & 18.1$\pm$39.0 & 62.2$\pm$24.4 \\
\hdashline
\checkmark & \checkmark & & & 75.4$\pm$23.0 & 36.7$\pm$67.1 & 11.6$\pm$25.4 & 64.8$\pm$23.1 \\
\checkmark$^{\dagger}$ & & & \checkmark& 70.2$\pm$24.4 & 37.2$\pm$74.7 & 11.7$\pm$25.7 & 58.4$\pm$23.4 \\
\checkmark$^{\ddagger}$ & & & \checkmark& 74.4$\pm$24.9 & 35.6$\pm$73.1 & 12.7$\pm$31.9 & 64.0$\pm$24.4 \\
\checkmark & \checkmark & \checkmark & & 75.9$\pm$23.4 & 31.7$\pm$68.1 & 9.5$\pm$23.2 & 65.4$\pm$23.2 \\
\hline
\checkmark & \checkmark & \checkmark & \checkmark & \textbf{76.5}$\pm$\textbf{22.5} &\textbf{28.9}$\pm$\textbf{64.7} & \textbf{8.7}$\pm$\textbf{22.1} & \textbf{66.1}$\pm$\textbf{22.7}\\
\specialrule{.8pt}{0pt}{2pt}
\end{tabular}
}
\label{tab:components}
\end{table}

\subsubsection{Analysis of Various Pseudo labels}
\label{aovps}
We conducted a quantitative evaluation of training the model directly using different graphical pseudo labels. Table \ref{tab:robuseness} lists the quantitative comparison based on the Dice score. The results indicated that the graphical pseudo labels exhibited a consistent performance ranking on both the Breast Ultrasound and Thyroid Ultrasound datasets. Specifically, the segmentation performance ranked in the following order from least to most favorable: Concavity, Circle, Quadrilateral, Box, Rectangle, and Irregular Ellipse.

{\color{black}
We categorized six geometric pseudo-labels into 'radical' and 'conservative' groups based on nodule morphology. Radical labels often over-segment, whereas conservative labels tend to under-segment relative to the ground truth. Our validation involved comparing these pseudo-labels, generated from aspect ratio annotations, with the ground truth, focusing on precision and recall. Our experiments, as shown in Table \ref{classify}, reveal that Circle, Box, Rectangle, and Irregular Ellipse (radical labels) have high recall, indicating over-segmentation, while Quadrilateral and Concavity (conservative labels) demonstrate high precision, indicating under-segmentation. These findings validate our classification approach.
}
\begin{table}[t!]
\small
\centering
\caption{\small{robustness analysis. we measure the corresponding dice score on tyroid ultrasound dataset and breast ultrasound dataset.}}
\renewcommand\arraystretch{1.4}
\newcolumntype{P}[1]{>{\centering\arraybackslash}p{#1}}
\resizebox{\linewidth}{!}{
	\begin{tabular}{ccccccc}
        \specialrule{.8pt}{0pt}{2pt}
        \multirow{2}{*}{\#} &
        \multirow{2}{*}{Pseudo Label} &
        \multicolumn{2}{c}{Thyroid Ultrasound}  && \multicolumn{2}{c}{Breast Ultrasound} \\ 
        \cline{3-4} \cline{6-7} 
        &&w/o $\mathcal{L}_{CAP}$ &with $\mathcal{L}_{CAP}$ &&w/o $\mathcal{L}_{CAP}$&with $\mathcal{L}_{CAP}$ \\
\specialrule{.4pt}{2pt}{0pt}
1&GrabCut~\cite{rother2004grabcut}& 68.6$\pm$25.3 & \textbf{71.5}$\pm$\textbf{25.1} &&71$\pm$23.0 & \textbf{71.9}$\pm$\textbf{21.5}\\
2&Quadrilateral& 67.7$\pm$27.2& \textbf{70.2}$\pm$\textbf{24.4}&&67.3$\pm$20.4& \textbf{68.9}$\pm$\textbf{21.5}\\
3&Circle&65.4$\pm$24.3 & \textbf{67.6}$\pm$\textbf{20.1}&&65.0$\pm$21.5 & \textbf{65.5}$\pm$\textbf{17.0}\\
4&Box&68.7$\pm$21.5 & \textbf{68.8}$\pm$\textbf{19.9}&&71.3$\pm$17.4 & \textbf{71.6}$\pm$\textbf{20.1}\\
5&Rectangle&69.2$\pm$22.2 & \textbf{70.2}$\pm$\textbf{22.8}&&71.5$\pm$18.5 & \textbf{73.5}$\pm$\textbf{21.1}\\
6&Irregular Ellipse&73.1$\pm$24.3 & \textbf{74.4}$\pm$\textbf{24.9}&&74.4$\pm$20.0 & \textbf{75.3}$\pm$\textbf{19.6}\\
7&Concavity&48.3$\pm$24.6 & \textbf{52.5}$\pm$\textbf{20.1}&&40.8$\pm$17.1 & \textbf{42.5}$\pm$\textbf{18.0}\\
\specialrule{.8pt}{0pt}{2pt}
\end{tabular}
}
\label{tab:robuseness}
\end{table}
\subsubsection{Superiority of IDMPS}
\label{idm}
The IDMPS module plays a critical role in our proposed method, as it introduces a novel dual-model supervision strategy. This strategy generates more accurate pseudo-labels to alleviate the issues of over-segmentation and under-segmentation caused by directly using conservative and radical labels for training. To validate the effectiveness of our approach, we conducted ablation experiments.
\begin{table}[t!]
\color{black}
\small
\centering
\caption{\small{Comparison of Precision and Recall Values for Six Pseudo-Labels Generated from Aspect Ratio Annotations Against the Ground Truth.}}
\renewcommand\arraystretch{1.4}
\newcolumntype{P}[1]{>{\centering\arraybackslash}p{#1}}
\resizebox{\linewidth}{!}{
	\begin{tabular}{ccccccc}
    \specialrule{.8pt}{0pt}{2pt}
    \multirow{2}{*}{Pseudo Label} &
    \multicolumn{2}{c}{Thyroid Ultrasound}  && \multicolumn{2}{c}{Breast Ultrasound} \\ 
    \cline{2-3} \cline{5-6} 
    &  Precision(\%) & Recall(\%) && Precision(\%) & Recall(\%) \\
    \specialrule{.4pt}{2pt}{0pt}
    \multicolumn{4}{l}{\textit{Radical labels:}} \\
    Circle & 62.2$\pm$12.6 & \textbf{99.4}$\pm$\textbf{1.1} && 54.7$\pm$12.8 & \textbf{99.2}$\pm$\textbf{1.4}  \\
    Box & 66.0$\pm$10.0 & \textbf{98.7}$\pm$\textbf{2.6} && 69.5$\pm$7.3 & \textbf{95.8}$\pm$\textbf{5.0}  \\
    Rectangle  & 67.0$\pm$6.8 & \textbf{99.8}$\pm$\textbf{0.8} && 67.7$\pm$5.9 & \textbf{98.8}$\pm$\textbf{1.9}  \\
    Irregular Ellipse  & 86.3$\pm$8.5 & \textbf{97.8}$\pm$\textbf{2.1} && 83.8$\pm$6.9 & \textbf{95.3}$\pm$\textbf{3.7}  \\\hline
    \multicolumn{4}{l}{\textit{Conservative labels:}} \\
    Quadrilateral & \textbf{97.7}$\pm$\textbf{6.9} & 74.2$\pm$4.6 && \textbf{96.5}$\pm$\textbf{4.7} & 71.1$\pm$5.5  \\
    Concavity & \textbf{97.9}$\pm$\textbf{7.4} & 29.7$\pm$3.5 && \textbf{99.0}$\pm$\textbf{3.3} & 31.0$\pm$3.0  \\
    \specialrule{.8pt}{0pt}{2pt}
    \end{tabular}
}
\label{classify}
\end{table}

Given the utilization of two sub-networks in our framework, we explored the effects of various supervision methods on the dual-model network. These approaches include: 1) Baseline, which  trains the model using the radical label; 2) Consistency Regularization (CR)\cite{dolz2021teach}, which directly encourages similarity between the two predictions; 3) Cross Pseudo Supervision (CPS)\cite{chen2021semi}, which employs one model's output as a hard pseudo-label to supervise the other model; 4) The proposed IDMPS approach, which dynamically blends the outputs of two models in regions with inconsistent conservative and radical labels, using the blended mask to train both models specifically in those areas; In order to evaluate the robustness and effectiveness of our proposed approach, we compared these approaches using various combinations of conservative and radical labels. Experimental results are shown in Figure~\ref{fig:IDMPS}. When the radical labels remained constant, the use of Quadrilateral as the conservative label yielded better results compared to Concavity. Conversely, when the conservative labels were kept constant, employing Irregular Ellipse as the radical label produced the best outcome. Our proposed method showed improved and consistent performance across various combinations of conservative and radical labels.

{\color{black} In the IDMPS module, we employed a dynamic random generation strategy, assigning $\beta$ a random value between 0 and 1 to improve the model's adaptability. Ablation studies were conducted to explore how the model performs with different $\beta$ values, and these findings are detailed in Table \ref{tab:beta}. We evaluated the model's effectiveness in three scenarios: using a single model without the IDMPS module, using the IDMPS module with $\beta$ fixed at 0.5, and using the IDMPS module with $\beta$ set to a random value. The results indicate that the IDMPS module achieves the best performance when $\beta$ is set to a random value.}

\subsubsection{Superiority of clinical anotomy prior loss} {
\begin{figure}[!t]
\setlength\tabcolsep{1.6pt}
    \centering
    \small
    \includegraphics[width=1.0\linewidth]{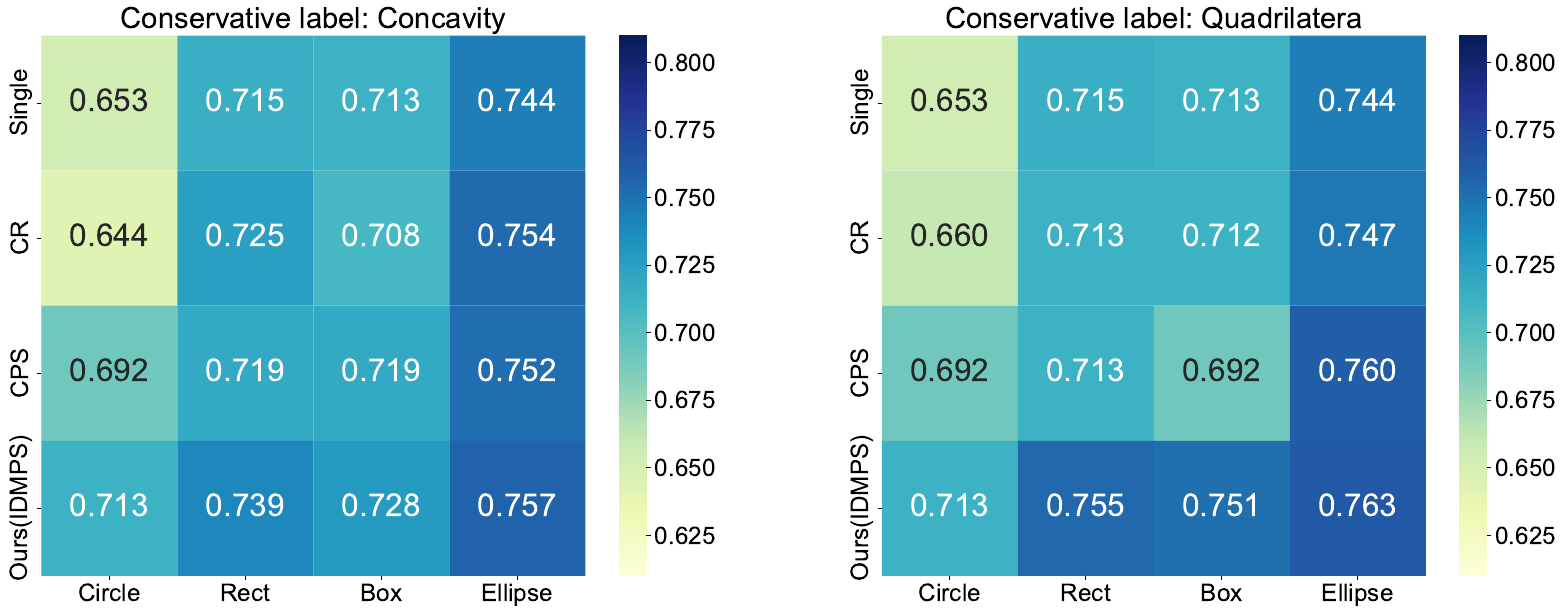}
    \caption{\small{Quantitative results of IDMPS with various combinations of conservative and radical labels. We measure the corresponding DSC on the breast ultrasound dataset. }}
    \label{fig:IDMPS}
\end{figure}
\begin{table}[t!]
\small
\color{black}
\centering
\caption{\small{Sensitivity analysis of $\beta$. We measure the corresponding DSC and Jaccard on thyroid ultrasound dataset. }}
\renewcommand\arraystretch{1.4}
\newcolumntype{P}[1]{>{\centering\arraybackslash}p{#1}}
\resizebox{\linewidth}{!}{
	\begin{tabular}{lcc}
		\toprule
		Method &DSC(\%) & Jaccard(\%) \\ \hline
		Single Model ($\mathcal{F}_{con}: \beta=0$) & 73.1$\pm$24.3 & 62.2$\pm$24.4 \\ 
            Single Model ($\mathcal{F}_{rad}: \beta=0$) & 67.7$\pm$27.2 & 56.3$\pm$25.2 \\ \hline
		IDMPS ($\beta=0.5$) & 74.7$\pm$23.4 & 63.8$\pm$23.0 \\
		IDMPS ($\beta=random$) & \textbf{75.4$\pm$23.0} &\textbf{64.8$\pm$23.1}\\ 
		\bottomrule
	\end{tabular}
}
\label{tab:beta}
\end{table}
The introduction of $\mathcal{L}_{CAP}$ aims to fully leverage the anatomical prior knowledge provided by the clinical annotation, including spatial location and the ratio of the long and short axes, to improve the model's accuracy in target localization and shape estimation. To evaluate the effectiveness of $\mathcal{L}_{CAP}$, we conducted an ablative experiment where the model was trained both with and without incorporating $\mathcal{L}_{CAP}$. During the training process, we utilized a variety of pseudo-labels, including conservative labels, radical labels, and labels generated by the GrabCut algorithm. Subsequently, we compared the performance of the model trained with $\mathcal{L}_{CAP}$ to that of the model trained without it. Table \ref{tab:robuseness} presents the quantitative results. The results demonstrated that regardless of whether the pseudo labels generated by GrabCut or conservative labels or radical labels were used for model training, incorporating $\mathcal{L}_{CAP}$ consistently improved the performance of the model.
}

\subsubsection{Hyper-parameters Experiments Results} {
The proposed framework includes a hyper-parameter $\alpha$ and two loss weights, $\lambda_{1}$ and $\lambda_{2}$. Firstly, we conducted experiments to investigate the impact of different values of the hyper-parameter $\alpha$ on our model. As shown in Figure \ref{fig:para}, the best performance was achieved when $\alpha$ was set to 3. Next, we designed experiments to examine the influence of different loss weight values, $\lambda_{1}$ and $\lambda_{2}$, on our model. We tested our method using different values of $\lambda_{1} \in \{0.1,0.3,0.5,1\}$ and $\lambda_{2} \in \{0.1,0.3,0.5,1\}$. Additionally, we attempted to determine the values of $\lambda_{1}$ and $\lambda_{2}$ using a Gaussian ramp-up function \cite{laine2016temporal}. This function adjusts the weights based on the total number of training epochs, allowing for a more adaptive approach in determining the appropriate weights. The summary results are shown in Figure \ref{fig:para}. The best result of our method was obtained when $\lambda_{1}$ was determined using a Gaussian ramp-up function and $\lambda_{2} = 0.3$.
}

\subsubsection{Comparison at Different Noise Levels of Aspect Ratio Annotations}
\begin{figure}[!t]
\setlength\tabcolsep{1.6pt}
    \centering
    \small
    \includegraphics[width=1.0\linewidth]{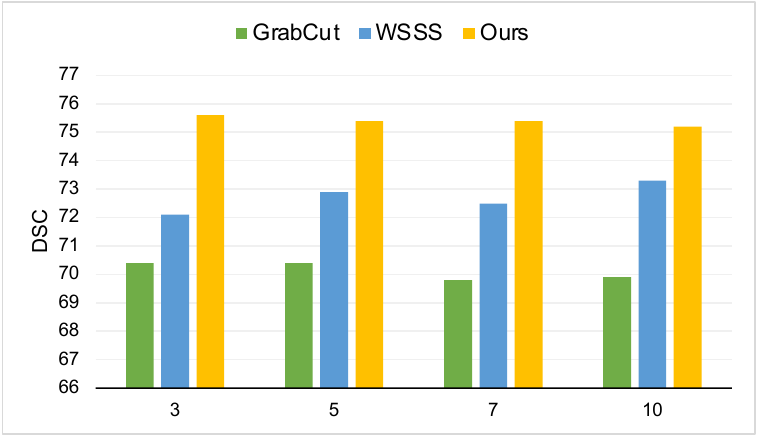}
    \caption{\small{Results of Noise Level Variations: We measured the corresponding DSC on the Thyroid Ultrasound dataset. Our method was compared with the baseline method and the second-place method. }}
    \label{fig:rotate}
\end{figure}

Due to variations in doctors' experience, the aspect ratio annotations of nodules made by doctors in clinical settings may contain noise. To further explore the segmentation performance at different noise levels, a series of experiments were conducted. Specifically, we introduced disturbances of $3^\circ$, $5^\circ$, $7^\circ$, and $10^\circ$ into the aspect ratio annotations. These perturbed aspect ratio annotations were then used for model training. We compared our proposed method with GrabCut~\cite{rother2004grabcut} and WSSS~\cite{cai2018accurate} (the second-place method) at different noise levels. The experimental results are presented in Figure \ref{fig:rotate}. The results show that as the noise in the aspect ratio annotations gradually increases, the performance of our proposed method declines. This is mainly because the introduction of noise causes a shift in the uncertain areas within the IDMPS module. Additionally, the performance of GrabCut and the GrabCut-based WSSS method varies more randomly. This is largely because the noise in the aspect ratio annotations has a minimal impact on the definition of the trimap required by GrabCut. However, our method outperforms WSSS at different noise levels and does not require a time-consuming iterative update strategy.

{\color{black}
\section{Discussion}
We develop a weakly supervised ultrasound nodule segmentation framework that achieves state-of-the-art (SOTA) performance. 
This framework is specifically designed to align with nodule characteristics, marking the first use of aspect ratio annotations for training in this context. Our results in Section \ref{aca} indicate that this approach significantly reduces the annotation effort compared to traditional methods. Beyond cost reduction, aspect ratio annotations are a standard clinical practice, streamlining the training process without extra annotation workload for clinicians.
Furthermore, we introduce a unique ultrasound nodule segmentation dataset, which includes aspect ratio and accurate manual annotations specifically for nodules in the breast and thyroid, establishing the first dataset of its kind in this domain. This resource is poised to be a valuable tool for researchers, potentially enhancing efficiency in clinical nodule segmentation.
\begin{table}[t!]
\small
\color{black}
\centering
\caption{\small{Performance of Various Modules in a Fully Supervised Setting.  We measure the corresponding DSC and Jaccard on breast ultrasound dataset. }}
\renewcommand\arraystretch{1.4}
\newcolumntype{P}[1]{>{\centering\arraybackslash}p{#1}}
\resizebox{\linewidth}{!}{
	\begin{tabular}{lcc}
		\toprule
		Method &DSC(\%) & Jaccard(\%) \\ \hline
		Full Supervision (single model) & 76.4$\pm$20.6 & 65.5$\pm$22.4 \\ \hline
		+IDMPS & 76.5$\pm$21.9 & 66.0$\pm$22.9 \\
		+IDMPS+CRB & 77.9$\pm$19.7 & 67.2$\pm$21.6\\
		+$\mathcal{L}_{CAP}$ & 77.7$\pm$20.2 & 67.1$\pm$21.8 \\\hline
            Ours (weakly supervision) & 76.6$\pm$19.5 & 65.3$\pm$20.5\\ 
		Ours (full supervision) & \textbf{78.5$\pm$18.5} & \textbf{67.7$\pm$20.6}\\ 
		\bottomrule
	\end{tabular}
}
\label{tab:fully}
\end{table}

In Section \ref{intro}, we categorize six basic geometric shape pseudo-labels into conservative and radical labels based on observations and the nodules' morphology. The classification of radical and conservative labels is based on their tendencies toward over-segmentation and under-segmentation, respectively, in comparison to the ground truth. Experiments in Section \ref{aovps}, using precision and recall metrics, validate our hypothesis. The results demonstrate that the irregular ellipse and quadrilateral are the most appropriately radical and conservative labels, respectively, for ultrasound nodule ground truth. Quantitative results from the IDMPS module in Section~\ref{idm} further corroborate this, showing that the combination of irregular ellipse and quadrilateral achieves superior segmentation performance. Therefore, the pseudo-label combination ultimately employed in our framework consists of irregular ellipse and quadrilateral. 
Despite the positive results from our proposed method, there are instances where specific lesions lead to the IDMPS uncertainty regions not fully encapsulating the lesion boundaries, which could potentially affect training outcomes. 
However, these scenarios are infrequent, as radical labels typically achieve high recall and conservative labels attain high precision, thereby limiting their impact on the overall training effectiveness.

In Section \ref{cwsota}, we show that our method outperforms a fully supervised UNet in a weakly supervised context, despite using two U-Net networks. To understand this better, we integrated our modules into a fully supervised setting, conducting ablation studies whose results are in Table \ref{tab:fully}. Our IDMPS module's comparison with CPS \cite{chen2021semi} and CR \cite{dolz2021teach} strategies in Section \ref{idm} revealed that while dual-model co-training reduces noise in weak supervision, it doesn't significantly boost performance in full supervision. Notably, our CRBS module significantly improved performance by introducing pseudo-label space perturbations through distinct misclassification costs. Additionally, our clinical anatomy prior loss was found to enhance performance in a fully supervised setting, with our framework reaching a DSC of 78.5 in full supervision and 76.6 in weak supervision.

Our implementation is designed for 2D ultrasound images, but it can be adapted for 3D spatial structures. To transition our IDMPS and CRBS modules to 3D, we would modify the models from 2D to 3D, such as transforming 2D UNet~\cite{unet2015MICCAI} into its 3D UNet \cite{cciccek20163d}. For the clinical anatomy prior loss, an adjustment to include an additional constraint along the z-axis would be necessary for 3D integration. The adaptation's main challenge is generating pseudo-labels for 3D. If every lesion location within a volume has an aspect ratio annotation, our method can generate pseudo-labels for each slice. If annotations are only available for some slices, a label propagation technique would be needed to extend annotations to the unannotated slices ~\cite{wang2022recistsup}. }

\section{Conclusion}
How to reduce the annotation cost remains a critical problem in ultrasound nodule segmentation. In this paper, we propose a novel approach for ultrasound nodule segmentation using clinical aspect ratio annotations. Compared to accurate annotations, aspect ratio annotations significantly reduce annotation costs. On the other hand, measuring the aspect ratio is a routine step in clinical diagnosis, which does not require additional workload for doctors. Our framework directly generates geometric pseudo-labels based on clinical annotations and differentiates them into conservative and radical labels according to their characteristics. We developed three components, namely CRBS, IDMPS and a clinical anatomy prior loss, to train our segmentation model. Experimental results demonstrated that our method outperforms other related methods, yielding superior segmentation accuracy. Additionally, in this paper, we introduce a new dataset that includes two types: thyroid ultrasound and breast ultrasound, along with precise annotations of nodules in the images and their aspect ratio annotations.

{\small
\bibliographystyle{IEEEtran}
\bibliography{ref}
}

\end{document}